\documentclass[conference]{IEEEtran}
\IEEEoverridecommandlockouts
\usepackage{multirow}
\usepackage{graphicx}
\usepackage{amsmath,amssymb,amsfonts}
\usepackage{algorithm,algorithmic}
\usepackage[algo2e,linesnumbered]{algorithm2e} 
\usepackage{subfigure}
\usepackage{graphicx}
\usepackage{textcomp}
\usepackage{xcolor}
\usepackage{mathtools}
\usepackage{bbm}
\usepackage{soul}
\usepackage{cite}
\usepackage{makecell}
\usepackage{booktabs}
\usepackage{mathtools}
\usepackage{enumitem}
\usepackage[linkcolor=black,
            anchorcolor=black,
            citecolor=black]{hyperref}
\def\BibTeX{{\rm B\kern-.05em{\sc i\kern-.025em b}\kern-.08em
T\kern-.1667em\lower.7ex\hbox{E}\kern-.125emX}}
\begin{document}

\title{Scaling SNNs Trained Using Equilibrium Propagation to Convolutional Architectures}

\author{\IEEEauthorblockN{Jiaqi Lin, Malyaban Bal, Abhronil Sengupta}
\IEEEauthorblockA{\textit{School of Electrical Engineering and Computer Science} \\
\textit{The Pennsylvania State University}\\
University Park, PA 16802, USA \\
Email: \{jkl6467,  mjb7906,  sengupta\}@psu.edu}
}
\maketitle


\begin{abstract}
Equilibrium Propagation (EP) is a biologically plausible local learning algorithm initially developed for convergent recurrent neural networks (RNNs), where weight updates rely solely on the connecting neuron states across two phases. The gradient calculations in EP have been shown to approximate the gradients computed by Backpropagation Through Time (BPTT) when an infinitesimally small nudge factor is used. This property makes EP a powerful candidate for training Spiking Neural Networks (SNNs), which are commonly trained by BPTT. However, in the spiking domain, previous studies on EP have been limited to architectures involving few linear layers.
In this work, for the first time we provide a formulation for training convolutional spiking convergent RNNs using EP, bridging the gap between spiking and non-spiking convergent RNNs. We demonstrate that for spiking convergent RNNs, there is a mismatch in the maximum pooling and its inverse operation, leading to inaccurate gradient estimation in EP. Substituting this with average pooling resolves this issue and enables accurate gradient estimation for spiking convergent RNNs. We also highlight the memory efficiency of EP compared to BPTT. In the regime of SNNs trained by EP, our experimental results indicate state-of-the-art performance on the MNIST and FashionMNIST datasets, with test errors of 0.97\% and 8.89\%, respectively. These results are comparable to those of convergent RNNs and SNNs trained by BPTT. These findings underscore EP as an optimal choice for on-chip training and a biologically-plausible method for computing error gradients.
\end{abstract}


\section{Introduction}
Equilibrium Propagation (EP) has surfaced as an efficient and biologically plausible learning framework for training energy-based neural architectures, namely convergent recurrent neural networks consisting of bidirectionally connected neurons \cite{scellier2017equilibrium}.
It offers an unified computational circuit during two phases of training through the minimization of an energy function. In contrast, Backpropagation (BP) separates the computational circuits into the forward and backward passes. In backward pass, the explicit calculation of error signal requires a circuit different from the forward pass, and is considered to be biologically implausible \cite{crick1989recent}. In EP, error signals are propagated from local perturbations on the last layer of neurons through backward connection to preceding layers. The implicit computation on errors features spatial locality that benefits neuromorphic computations \cite{martin2021eqspike}. 
Theoretical analysis of EP \cite{scellier2019equivalence} has been shown to approximate Backpropagation Through Time (BPTT) \cite{almeida1990learning, pineda1987generalization}, and the weight updates of EP resembles Spike-Timing-Dependent Plasticity (STDP) \cite{bi1998synaptic, scellier2017equilibrium}, which advocates EP as an outstanding biologically plausible alternative to BPTT. However, EP suffers from the first-order bias in the gradient estimate due to finite nudging. Earlier works on EP, therefore were restricted to shallow fully-connected architectures \cite{scellier2017equilibrium, o2019training, ernoult2019updates, ernoult2020equilibrium}. To overcome this limitation, researchers incorporated both positive and negative nudge phases to cancel out the bias introduced, subsequently allowing for a more accurate gradient estimation. Recent advancements \cite{laborieux2021scaling, bal2023equilibrium} have reduced the accuracy disparity between EP and BP on both vision-based datasets and natural language processing based sequential learning tasks.
 
Spiking Neural Networks (SNNs) are considered to be a more biologically plausible neural model where the information communicated between neurons depends on spiking events. The event-driven mechanism significantly reduces the energy consumption of neural networks due to the sparsity of neuron activities \cite{mead1990neuromorphic, merolla2014million, benjamin2014neurogrid}. 
The majority of current works for training SNNs from scratch uses BP based methods \cite{neftci2019surrogate, he2020comparing,  yin2021accurate, bal2024spikingbert, fang2021deep, zhou2023spikformer}, which reduces the biological significance of the former. EP leverages unified circuitry and inherent dynamics of neural systems to adjust synaptic weights, which is analogous to how learning might occur in biological systems. The biological plausibility of Equilibrium Propagation (EP), coupled with its recent strides in narrowing accuracy gaps, indicates its potential to serve as a viable alternative for training SNNs. A recent study proposed a formulation of EP with a step-size scheduler that adjusts the accumulation rate of neurons states, allowing SNNs to converge to the same stable states as convergent RNNs \cite{o2019training}. However, there is a performance gap between this work and state-of-art performance on the MNIST dataset. 
A simpler yet effective methodology of training SNNs using EP has been also proposed with the only requirement of leaky integrate-and-fire (LIF) neurons \cite{martin2021eqspike} in the network formulation. Nonetheless, this simple EP system experiences scalability issues. Experimentally, extending this formulation into multiple linear layers results in vital performance degradation. Both studies on SNNs are confined to linear layers and only the MNIST dataset. Given the effectiveness of EP in training continuous-valued convolutional neural networks \cite{ernoult2019updates,laborieux2021scaling}, it is imperative to develop an approach to train convolutional spiking convergent RNNs using EP. However, convolutional architecture formulations from the continuous-valued domain cannot be directly applied to spiking convergent RNNs and therefore needs significant rethinking in architectural modules like maximum pooling and its inverse, namely unpooling operations.
The primary contributions of this work are the following: 
\begin{itemize}
    \item We explore for the first time how convolutional spiking convergent RNNs can be trained using EP, bridging the gap between spiking and non-spiking convergent RNNs in the regime of EP.
    \item We conduct theoretical and experimental analysis on the maximum pooling and unpooling operations to show the information misalignment between forward connections and backward connections. This issue is unique to spiking convergent RNNs trained by EP, which significantly degrades the performance of EP.
    \item We propose to implement EP by replacing maximum pooling and unpooling with average pooling and nearest neighbor upsampling, which solves the aforementioned problem. The use of average pooling along with upsampling enables us to achieve 0.97\% and 8.89\% test error on MNIST and FashionMNIST datasets, respectively.
\end{itemize}

\section{Methods\label{mtd}}

In this section, we first formulate EP by introducing the Hopfield network in terms of convergent RNNs. Then, we revisit methods to quantize EP based on spiking events such that spiking events match the neuron states in convergent RNNs. We subsequently propose convolutional operations in spiking convergent RNNs using EP, and further investigate the information loss in maximum pooling and unpooling operations in the spiking domain.  

\subsection{Hopfield Network\label{mtd:hn}}
A continuous-valued Hopfield Network is a network composed of recurrently connected neurons featuring symmetrical weights ($w_{ij} = w_{ji}$). The energy function $E(\cdot)$ for this type of network is defined as follows \cite{hopfield1984neurons}:
\begin{equation}
E(s) = \frac{1}{2}\sum\limits_{i} \xi^2_i - \frac{1}{2}\sum\limits_{i\not=j} w_{ij}\rho(\xi_i)\rho(\xi_j) - \sum\limits_{i}b_{i}\rho(\xi_{i})
\end{equation}
where, $\xi_i$ is the state of neuron $i$, $w_{ij}$ and $b_i$ are weights and biases of the neuron, and $\rho(\cdot) = [\cdot]^1_0$ is a nonlinear hard sigmoid function that clips the activation of neurons in a range between 0 and 1 \cite{scellier2017equilibrium}. 
The state transition dynamics of convergent RNNs are derived from this energy function. This imposes an inherent requirement on convergent RNNs, necessitating symmetric bidirectional connections across layers. In convergent RNNs, the temporal dynamics of the neurons is defined as:
\begin{equation}
\frac{\partial \xi^i}{\partial t} = - \frac{\partial E(\xi^i)}{\partial \xi^i} = -\xi^i + \rho^\prime(\xi^i)\left(\sum\limits_{j} w_{ij}\rho(\xi^j) + b_{i}\right)
\end{equation}
Based on the Euler Method, the discretization of this formula gives:
\begin{equation}
\label{nd1}
\xi^t_i = \rho\left((1-\epsilon)\xi^{t-1}_i + \epsilon\rho^\prime(\xi^{t-1}_i)\left(\sum\limits_{j} w_{ij}\rho(\xi^{t-1}_j) + b_{i}\right)\right)
\end{equation}
where, $\epsilon$ is the step-size for updating the neuron states. These dynamics allows the neuron to converge to a stable state. 

\subsection{Equilibrium Propagation}
EP, a two-phase learning algorithm, was initially proposed for training continuous-valued Hopfield Networks \cite{scellier2017equilibrium} and later extended to implement gradient descent in convergent RNNs on a loss function $L(\xi^{\text{out}}, y)$ between some target $y$ and output activation $\xi^{\text{out}}$. In the first phase, named as the free phase, a static input $x$ is fed into the neural network over $T_{\text{free}}$ time steps until the state of the network is saturated to a state $\xi^*$ with minimized energy $E(\xi^*)$. The second phase is a nudge phase with a nudging term $\beta \frac{\partial L(\xi^{\text{out}}, y)}{\partial \xi}$, where the parameter $\beta$ is the nudging factor. The nudging term allows the states of the neural network to settle to a stable state $\xi^{\beta}$ in $T_{\text{nudge}}$ time steps. The parameter updates are based on the divergence of stable states in the two phases. With $\beta \rightarrow 0$, the gradients of EP calculated below approximate the gradient calculated by BPTT \cite{scellier2019equivalence}:
\begin{equation}
    \label{ep1}
\Delta w = \frac{1}{\beta} \left( \frac{\partial E(\xi^{\beta})}{\partial w} - \frac{\partial E(\xi^{*})}{\partial w}\right)
\end{equation}
Laborieux \textit{et al.} observed the estimator bias introduced in the nudge phase with a positive value of $\beta$, leading to an inaccurate estimate of gradients \cite{laborieux2021scaling}. To address this issue, a third phase with a nudging factor of $-\beta$ is introduced such that the estimation biases in the second phase and third phase are canceled out. Equation \ref{ep1} therefore becomes:
\begin{equation}
\label{ep2}
\Delta w = \frac{1}{2\beta} \left( \frac{\partial E(\xi^{\beta})}{\partial w} - \frac{\partial E(\xi^{-\beta})}{\partial w}\right)
\end{equation}

\subsection{Spiking Module}
Following the energy-based neuron dynamics in Equation \ref{ep2}, O’Connor \textit{et al.} proposed a method that allows neurons with binary communication to converge to the same fixed point as convergent RNNs using EP \cite{o2019training}. In this formulation, the Sigma-Delta modulation $\delta(\cdot)$ \cite{candy1991oversampling} mimics the subtractive LIF mechanism of spiking neurons, which is defined as:
\begin{equation}
\begin{split}
s^t &= \sigma(V^{t-1} + x^t > V_{th}) \\
V^t &= V^{t-1} + x^t - s^t
\end{split}
\end{equation}
where, at time step $t$, $V^t$ is the membrane potential, $x^t$ is the input, $s^t_i$ is the spike generated at time $t$, and $\sigma(\cdot)$ is the activation function that generates a spike when the membrane potential exceeds the threshold $V_{th}$. To accelerate binary communication between neurons, predictive coding is exploited to encode and decode real-valued information into bit-streams \cite{o2019training}. In this scheme, to leverage the bandwidth of communication, only temporal changes in neuron states is sent through binary bit-streams to succeeding neurons via predictive encoder. Subsequently, the predictive decoder receive these changes and estimates current signal valued $d^t_i$ as a linear function of previous samples $d^{t-1}_i$, where $\lambda$ is the prediction factor:
\begin{equation}
d^t_i = (1-\lambda)d^{t-1}_i + \lambda\sum\limits_j w_{ij}s^{t-1}_j
\end{equation}
The predictive encoder quantizes temporal changes in neuron states into binary information, where $e^t_i$ is predicted encoding signal accumulated until time step $t$:
\begin{equation}
\begin{split}
e^t_i &= \frac{1}{\lambda}(\rho(\xi^t_i) - (1-\lambda)\rho(\xi^{t-1}_i) \\
s^t_i &= \delta(e^t_i)
\end{split}
\end{equation}

\subsection{Spiking Convolutional Layer}
Motivated from convolutional architectures for non-spiking convergent RNNs with static input \cite{ernoult2019updates, laborieux2021scaling}, in this paper we propose spiking convolutional layer in the context of energy-based systems. Given $N_{\text{c}}$ number of convolutional layers followed by $N_{\text{l}}$ number of linear layers, input function $\phi(s,t,n)$ sums up weighted quantized inputs of layer $n$ at time $t$ accommodating spike-based communications:
\begin{equation}
    \phi(s,t,n) =
    \begin{cases}
    \mathcal{P}(w_n \ast s^{t-1}_{n-1}) + w_{n+1} \Tilde{\ast} \hspace{1px} \mathcal{P}^{-1}(s^{t-1}_{n+1}), & \\
    \qquad \text{if} \quad 1\leq n\leq N_{\text{c}} &\\
    w_n \cdot s^{t-1}_{n-1} +w^T_{n+1} \cdot s^{t-1}_{n+1}, & \\
    \qquad \text{if} \quad N_{\text{c}} < n \leq N_{\text{t}} &
    \end{cases}
\end{equation}
where, $N_{\text{t}} = N_{\text{c}} + N_{\text{l}}$, $\mathcal{P}(\cdot)$ is the pooling operation, $\mathcal{P}^{-1}(\cdot)$ is the inverse of corresponding pooling operation, $w_n$ is the parameter of layer $n$. To distinguish the operations in linear layers and convolutional layers, let $\ast$ be the convolution operation, $\Tilde{\ast}$ be the convolution transpose operation, and $\cdot$ be the linear mapping operation.
Connecting to Sigma-Delta modulation and predictive coding, the neuron dynamics is formulated as:
\begin{equation}
\label{eq:snn_nd}
\begin{split}
d^t_n &= (1-\lambda)d^{t-1}_n + \lambda 
\phi(s,t,n) \\
\xi^t_n &= \rho\left((1-\epsilon)\xi^{t-1}_n + \epsilon\rho^\prime(\xi^{t-1}_n)\left(d^t_n + b_{n}\right)\right)\\
e^t_n &= \frac{1}{\lambda}(\rho(\xi^t_n) - (1-\lambda)\rho(\xi^{t-1}_n) \\
s^t_n &= \sigma(V^{t-1}_n + e^t_n > V_{th}) \\
V^t_n &= V^{t-1}_n + e^t_n - s^t_n
\end{split}
\end{equation}
To leverage spatial locality, weight updates are directly derived from the states of the two connecting layers $n$ and $n+1$. Accommodating both linear and convolutional layers, Equation \ref{ep2} becomes:
\begin{equation}
\Delta w_n =
    \begin{cases}
         \frac{1}{2\beta} \left(\mathcal{P}^{-1}(\xi^{\beta}_{n+1}) \ast  \xi^{\beta}_{n} - \mathcal{P}^{-1}(\xi^{-\beta}_{n+1}) \ast  \xi^{-\beta}_{n}  \right), &\\ 
         \qquad \text{if} \quad 1\leq n\leq N_{\text{c}} &\\
         \frac{1}{2\beta} \left( {\xi^{\beta}_{n+1}}\cdot{\xi^{\beta}_{n}}^T - {\xi^{-\beta}_{n+1}}\cdot{\xi^{-\beta}_{n}}^T\right), &\\
         \qquad \text{if} \quad N_{\text{c}} < n \leq N_{\text{t}} &
    
    \end{cases}
\end{equation}

\subsection{Rethinking Pooling and Unpooling Operations}
In this section, theoretical and experimental analysis is provided to justify the reformulation of pooling and unpooling operations in the energy-based spiking convolutional convergent RNN architecture.
\begin{figure}[b!]
  \centering
  \includegraphics[scale=0.3]{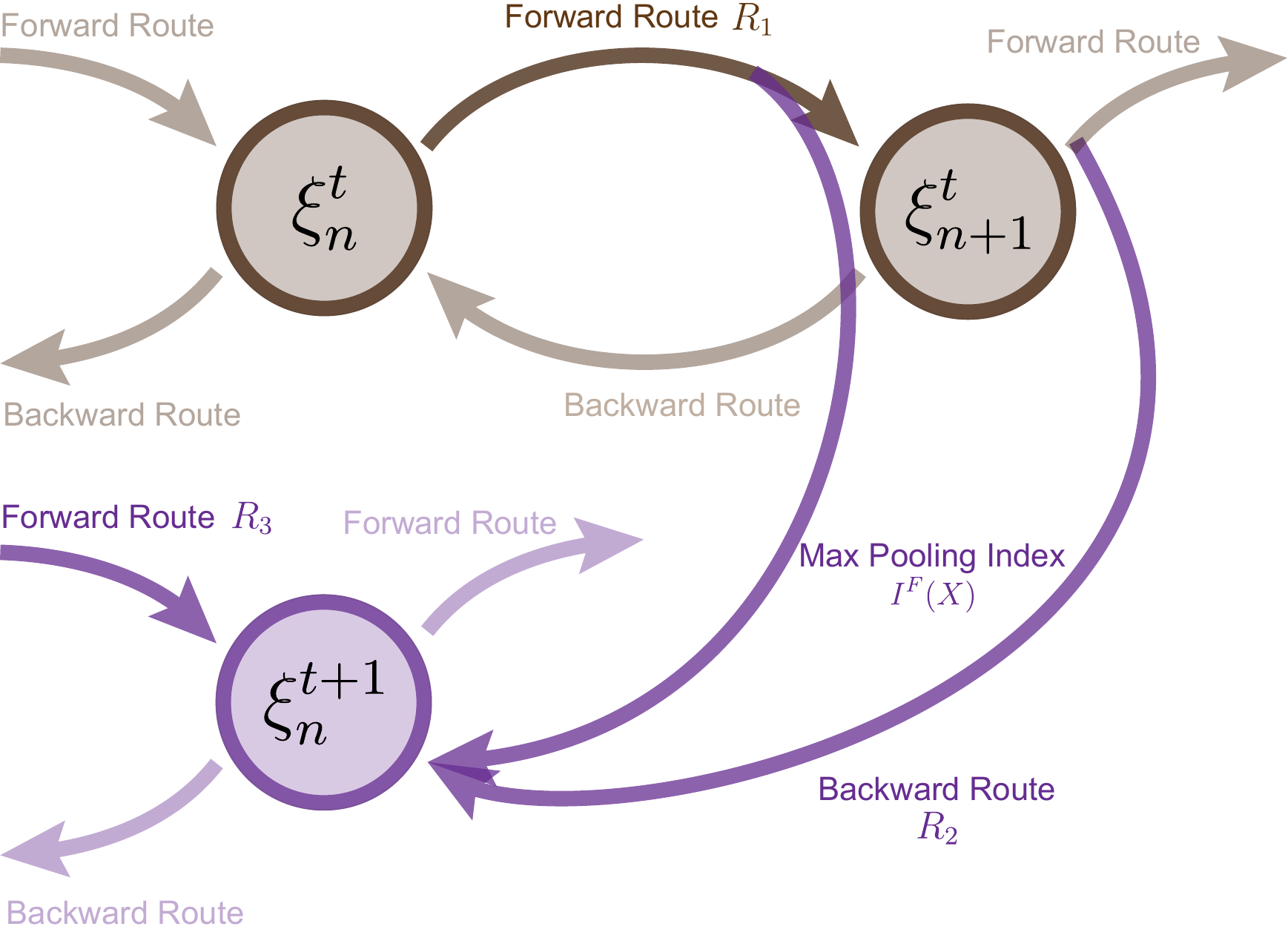}
  \caption{Information transmission and retention between two consecutive neurons during a period of 2 time steps. Brown color represents information at $t$, and purple color represents information at $t+1$.}
  \label{fig:maxpool}
\end{figure}


\begin{figure*}
  \centering
  \includegraphics[scale=0.55]{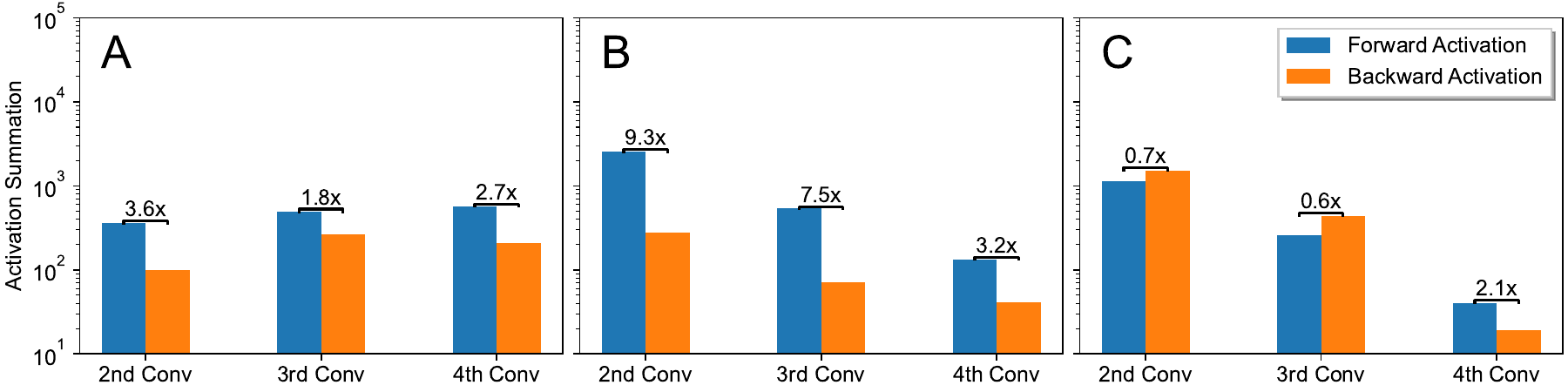}
    \caption{Activation summation of middle 3 convolutional layers with 32-64-128 channels in both forward route and backward route during nudge phase of a five-layer convolutional architecture. The activations are mean over 2000 random training samples from the MNIST dataset and summed across spatial dimensions of the convolutional layers. (A) Activation summation of convergent non-spiking RNNs equipped with maximum pooling and unpooling operations; (B) Activation summation of SNNs equipped with maximum pooling and unpooling operations; (C) Activation summation of SNNs equipped with average pooling and its inverse operator.}
  \label{fig:act}
\end{figure*}

Let us first explain the pooling indices mismatch problem in maximum pooling and unpooling operations. Figure \ref{fig:maxpool} illustrates a toy example of information transmission in EP between two consecutive neurons at time steps $t$ and $t+1$. In forward route, $R_1$ from $\xi^t_n$ to $\xi^t_{n+1}$, a pooling operator is applied after the convolution of spikes quantized from $\xi^t_n$, returning down-sampled continuous values with corresponding indices $I^{F}(X)$. $I^{F}(X)$ represents the position of the maximum value in each pooling zone, $F$ is the filter size of maximum pooling operator, and $X$ is the input of the pooling operation. Information output from $\xi^t_{n+1}$ is then combined with $I^{F}(X)$ and subsequently passed to $\xi^{t+1}_n$ through backward connection $R_2$. $I^{F}(X)$ is used as the positional indicator to upsample the information implicitly conveying error signal through the inverse of maximum pooling operation. 
A maximum pooling operation $\mathcal{P}_{\text{max}}(\cdot)$ is defined as \cite{ernoult2019updates}:
\begin{equation}
    \label{eq:maxpool}
\mathcal{P}_{\text{max}}(X, F)_{c,i,j} = \max_{p,q\in[0, F-1]}\left\{ X_{c,F (i-1)+1+p,F (j-1)+1+q} \right\}
\end{equation}
where, $c$ is the channel, $i,j$ is the spatial position of input $X$, and $p,q$ are the spatial indices in the pooling zone. The relative indexing $I^{F}(X)$, computed by Equation \ref{eq:maxpool} within a pooling zone, is defined as:
\begin{equation}
    I^{F}(X)_{c,i,j} = \underset{p,q\in[0, F-1]}{\text{arg\,max}}\left\{ X_{c,F (i-1)+1+p,F (j-1)+1+q} \right\}
\end{equation}
Consider $\mathcal{P}_{\text{max}}(X, F)$ gives $Y$ and $I^F(X)$. The inverse of maximum pooling operating on $Y$ is defined as:
\begin{equation}
\mathcal{P}^{-1}_{\text{max}}(Y, I^F(X))_{c,i,j} =
    \begin{cases}
        Y_{c, \lceil i/F\rceil, \lceil j/F\rceil}
            & \text{if $\{i,j\}\in I^{F}(X)$} \\
        0 & \text{otherwise} \\
    \end{cases}
\end{equation}
In convergent RNNs, $Y$ returned by $\mathcal{P}_{\text{max}}(X, F)$ forms a one-to-one relationship with indices $I^{F}(X)$ in forward route $R_1$, where $X= w_n \ast \xi^{t}_{n}$. The information in $Y$ is retained in the evolution of neuron states $\xi^{t}_{n+1}$, and is returned in backward route $R_2$ without the loss of information. However, in the backward route $R_2$ of the spiking implementation, $Y$ returned by $\mathcal{P}_{\text{max}}(X, F)$ is quantized through Sigma-Delta modulation and predictive coding, as described in Equation \ref{eq:snn_nd}. During the process of updating neuron states $\xi^{t}_{n+1}$, $Y$ is partially quantized into binary spikes $s^{t}_{n+1}$ and partially accumulated as potential $V^{t}_{n+1}$, thereby resulting in a possibility of information mismatch with $I^{F}(X)$. Consequently, $I^{F}(X)$ cannot be used as a positional indicator for the inverse of maximum pooling in backward route $R_2$. 
An alternative solution is to perform maximum pooling after the quantization step. However, this approach leads to inconsistencies in the spatial dimensions between the forward and backward routes, entailing the use of additional upsampling operators after the inverse of maximum pooling. Specifically, the integration of forward route $R_3$ and backward route $R_2$ at neuron state $\xi^{t+1}_{n}$ necessitates that both inputs have identical spatial dimensions. In the information propagation route, signals at time $t$ pass through two neuron layers, undergoing downsampling twice after the quantization of each neuron state $\xi^{t}_n$ and $\xi^{t}_{n+1}$, and subsequently serve as backward input through $R_2$ to $\xi^{t+1}_n$, where signals are upsampled once. Before integration, the forward input does not undergo any downsampling or upsampling in $R_3$. Consequently, the backward input requires an additional upsampling step to align with the size of the forward input.

\begin{figure*}
  \centering
  \includegraphics[scale=0.55]{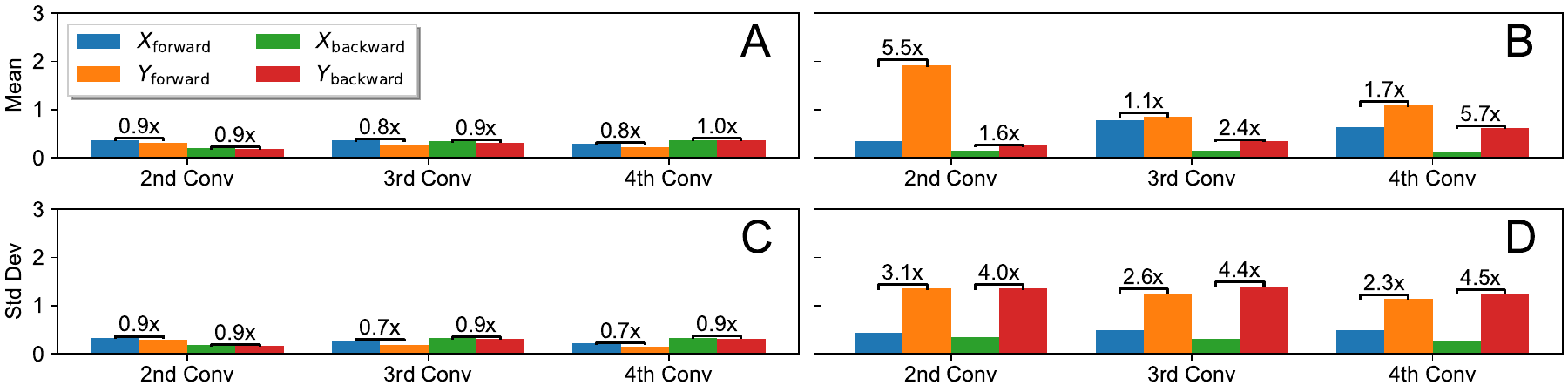}
  
      \caption{\label{fig:mean_std}Mean and standard deviation (Std Dev) of activations $X_{\text{forward}}$, $Y_{\text{forward}}$, $X_{\text{backward}}$, and $Y_{\text{backward}}$ of middle 3 convolutional layers with 32-64-128 channels during nudge phase of a five-layer convolutional architecture equipped with maximum pooling and unpooling operations. The average and standard deviation is calculated over 2000 samples from the MNIST dataset. (A) Mean activation of convergent RNNs; (B) Mean activation of SNNs; (C) Standard deviation of convergent RNNs' activations; (D) Standard deviation of SNNs' activations.}
\end{figure*}

Additionally, we show that maximum pooling and unpooling operations creates a significant activation magnitude imbalance between the forward and backward routes in the spiking regime. 
Experimental results in Figure. \ref{fig:act} presents the activation outputs of the maximum pooling operation and its inverse from both the forward and backward routes during the nudge phase. (The first convolutional layer is excluded since it receives a quantized static image as input in the forward route, making it inappropriate for comparison.) It reveals that magnitude differences in convergent RNNs equipped with maximum pooling and unpooling are less than $4\times$, while a significant gap in activation magnitudes is observed between the forward and backward routes, with differences ranging up to $10\times$ in the spiking regime. This discrepancy is more pronounced in shallower convolutional layers. 
To investigate why this imbalance is specific to spiking convergent RNNs, consider forward route $R_3$ and backward route $R_2$ in Figure. \ref{fig:maxpool}, and let $Y_{\text{forward}} = \mathcal{P}(w_n \ast X_{\text{forward}})$ and $ Y_{\text{backward}} = w_{n+1} \Tilde{\ast} \hspace{1px} \mathcal{P}^{-1}(X_{\text{backward}})$, where $X_{\text{forward}}$ and $X_{\text{backward}}$ are the inputs of forward routes $R_3$ and backward route $R_2$, respectively. Figure. \ref{fig:mean_std} illustrates an increase of both mean and standard deviation from $X_{\text{forward}}$ to $Y_{\text{forward}}$ and from $X_{\text{backward}}$ to $Y_{\text{backward}}$ in the spiking domain. Conversely, continuous-valued convergent RNNs exhibit only slight increment. This significant magnitude difference in the spiking scenario is due to the quantization of continuous-valued neuron states to binary spike values, where binary discretization to 0 and 1 values can significantly deviate from the original information and magnify minor differences.
In EP, error signal propagation depends on backward routes. The imbalance problem trivializes the significance of the error signal during the accumulation of potentials, resulting in an ineffective nudge phase update of neuron states.

Applying a maximum pooling and its inverse operation in spiking convergent RNNs therefore hinders the accurate estimation of EP. To circumvent these issues, we propose to replace all the maximum pooling and unpooling operations with average pooling and use nearest neighbor unsampling for the inverse of pooling operation. Average pooling and its inverse does not necessitate indices, which avoids the pooling indices mismatch problem. It comes with the advantages of simpler circuit design in neuromorphic hardware, as storing indices require additional states for information retention, and the transmission necessitates extra resources in addition to spike communication. Additionally, the nearest neighbor upsampling method copies the pixel value from the non-zero input feature map to all positions in a pooling zone, creating a pooling zone without zeros to augment the information passing backwards. This upsampling method mitigates the activation magnitude imbalance problem observed in maximum pooling and its inverse operators.
We define the average pooling operations as:
\begin{equation}
        \mathcal{P}_{\text{avg}}(X, F)_{c,i,j} = \frac{1}{F^2}\sum_{p,q\in[0, F-1]} X_{c,F (i-1)+1+p,F (j-1)+1+q}
\end{equation}
We introduce the inverse of average pooling as:
\begin{equation}
\mathcal{P}^{-1}_{\text{avg}}(Y)_{c,i,j} = \frac{Y_{c, \lceil i/F\rceil, \lceil j/F\rceil}}{\alpha}
\end{equation}
where the scaling factor $\alpha$ controls the magnitude of backward propagating signal. Figure. \ref{fig:act}.C presents the activations of average pooling operations from both the forward and backward routes during the nudge phase with $\alpha=F^2$ (chosen as an optimal setting for our experiments and can be tuned further). This demonstrates that average pooling mitigates the magnitude imbalance between the two routes. The experimental performance of our proposed modifications are presented in the following section.

\section{Experimental Results}

In this section, EP is implemented using the methodology outlined in Section \ref{mtd}. Spiking convergent RNNs are utilized for visual classification tasks on the MNIST and FashionMNIST datasets. For comparison, the performance of SNNs trained by BPTT and convergent RNNs trained using both EP and BPTT are also evaluated, denoted as SNN BPTT, CRNN EP, and CRNN BPTT, respectively. With the proposed modifications, we achieve comparable performance on these visual tasks to those of neural networks trained via BPTT. All experiments using BPTT employ the same corresponding architecture as those in EP. Stochastic Gradient Descent (SGD) without momentum is implemented. Additionally, Mean Squared Error (MSE) is applied to spiking convergent RNNs and SNNs, and Cross-Entropy Loss (CE) is applied to convergent RNNs. Optimal hyperparameters are applied to achieve the best performance in each experiment for both EP and BPTT frameworks.

\begin{table}[t]
\centering
  \caption{Hyper-parameters for MNIST and FashionMNIST Datasets.}
  \label{tab:param}
  \begin{tabular}{lccc}  
  \toprule
     Hyper-parameters & Range & MNIST & FashionMNIST\\
    \midrule
     $\beta$ & (0.01-1.0) & 0.1 &  0.2 \\
     $\epsilon$ & (0.1-1.0) & 0.9 &  1.0 \\
     $\lambda$ & (0.1-1.0) & 0.6 &  0.6 \\
     $T_{\text{free}}$ & (50-500) & 250 &  350 \\
     $T_{\text{nudge}}$ & (10-100) & 50 &  50\\
     \multirow{2}{*}{Learning Rate$^*$} & \multirow{2}{*}{(0-1.0)} &[0.25, 0.15, & [0.25, 0.15, \\
      &  & 0.1, 0.08] & 0.1, 0.08]\\
     Batch Size & (10-500)&125 & 125 \\
     Epochs & (10-1000) & 250 & 250 \\
    \bottomrule
    \multicolumn{4}{l}{\parbox [t] {210pt}{$^*$A layer-wise learning rate is implemented.}} \\
  
\end{tabular}
\end{table}

\begin{table*}[t]
\centering
  \caption{Error and Memory Consumption of EP and BPTT for Both Spiking and Non-Spiking Networks.}
  \label{tab:perf}
  \begin{tabular}{lcccccccccc}
  \toprule
   &  & \multicolumn{3}{c}{SNN EP Error (\%)}&\multicolumn{2}{c}{SNN BPTT Error (\%)} &\multicolumn{2}{c}{CRNN EP Error (\%)} &\multicolumn{2}{c}{CRNN BPTT Error (\%)} \\
   \cmidrule(lr){3-5} \cmidrule(lr){6-7} \cmidrule(lr){8-9} \cmidrule(lr){10-11} 
     Dataset&Model&Test&Train& MEM & Test& MEM &Test& MEM &Test & MEM\\
    \midrule
     \multirow{4}{*}{MNIST} & 1FC\cite{o2019training} & 2.37 &  0.15 & - &  - & - & - & - & - & - \\
     & 1FC\cite{martin2021eqspike} & 2.41& 1.09 & - & - & - & - & - & - & - \\
     & 3FC\cite{o2019training} &  2.42 & 0.27 & - & -&-  & -& -& -& -\\
     & 2C1FC\cite{ernoult2019updates} &  1.02 & 0.54 & - & -&-  & -& -& -& -\\
         & 2C2FC & $0.97\pm0.32$ & $0.61\pm0.28$ & 321 & $0.86\pm0.11$ & 2171 & $0.94\pm0.02$ & 309 & $0.99\pm0.36$&  3018  \\
     \hline
    {FashionMNIST}   &2C2FC  & $8.89\pm2.45$ & $2.90\pm0.02$ &  321 & $8.71\pm1.25$ &   2171 & $8.99\pm2.89$ & 309  & $8.83\pm1.33$&   3018 \\
  \bottomrule
\end{tabular}
\end{table*}

The employed convolutional architecture  consists of 2 convolutional layers and 2 linear layers and is denoted as 2C2FC. The convolutional layers consists of 128 and 256 channels, respectively, each of which has a kernel size of 3, stride of 1, and padding of 0. Each convolutional layer is followed by a 2-by-2 Average Pooling operation with a stride of two. The weights are initialized using the uniform Kaiming initialization \cite{he2015delving}. 
The range for tuning hyper-parameters is reported in Table \ref{tab:param}. The experiments were run on one Nvidia RTX 2080 Ti GPU with 11GB memory. Memory consumption is averaged across all batches within a single epoch, with the batch size fixed at 10. Memory consumption of BPTT and EP is averaged over 150 time steps. In EP, free phase, positive nudge phase, and negative nudge phase have 100, 25, and 25 time steps respectively. Training and testing error are averaged over 5 independent trials.

Table \ref{tab:perf} presents the performance of SNNs, and convergent RNNs trained by EP and BPTT. In the table, \#FC represents the number of linear layers, \#C denotes the number of convolutional layers, and MEM indicates the memory consumption measured in Megabytes. Spiking convergent RNNs trained by EP with the proposed modifications achieve performance at par with SNNs trained by BPTT and convergent RNNs trained by BPTT and EP, while they consume notable reduction in memory footprint compared to those trained by BPTT. 

\section{Discussion}
The EP learning framework regards underlying neural networks as energy-based dynamical systems, providing deeper insights into learning frameworks and serving as a catalyst for further exploration into cognitive processes within the brain. This study is the first exploration of employing EP as a learning mechanism within spiking convergent RNNs incorporating convolution operations, aimed at addressing the scaling challenges of EP to complex visual classification tasks. The performance of EP is limited by indexing mismatch issues caused by maximum pooling and unpooling operations, which are unique to the spiking domain. Substituting maximum pooling and its inverse with average pooling along with nearest neighbor unsampling has proven effective for EP learning. The local weight update feature of EP, both spatially and potentially temporally \cite{ernoult2020equilibrium}, eliminates the need to store computational graphs for weight updates, significantly reducing the memory consumption during training compared to BPTT. 
Additionally, it does not rely on surrogate gradients to address the non-differentiable nature of spiking neurons \cite{neftci2019surrogate, yin2021accurate, zenke2021remarkable}. Coupled with its inherent single-circuit design and STDP-like weight update mechanism \cite{scellier2017equilibrium}, EP is emerging as an optimal choice for on-chip training. It not only aligns closely with biological plausibility but also delivers comparable accuracy to iso-architecture ANNs. 
Future work can explore incorporation of maximum pooling operation in the spiking convergent RNN architecture by overcoming the aforementioned problems along with other innovations in the training process to close the performance gap between EP trained models and state-of-art neural networks trained by BP.


\section*{Acknowledgments}
This material is based upon work supported in part by the U.S. National Science Foundation under award No. CAREER \#2337646, CCSS \#2333881, CCF \#1955815, and EFRI BRAID \#2318101 and by Oracle Cloud credits and related resources provided by the Oracle for Research program.

\begin{thebibliography}{10}
\providecommand{\url}[1]{#1}
\csname url@samestyle\endcsname
\providecommand{\newblock}{\relax}
\providecommand{\bibinfo}[2]{#2}
\providecommand{\BIBentrySTDinterwordspacing}{\spaceskip=0pt\relax}
\providecommand{\BIBentryALTinterwordstretchfactor}{4}
\providecommand{\BIBentryALTinterwordspacing}{\spaceskip=\fontdimen2\font plus
\BIBentryALTinterwordstretchfactor\fontdimen3\font minus \fontdimen4\font\relax}
\providecommand{\BIBforeignlanguage}[2]{{%
\expandafter\ifx\csname l@#1\endcsname\relax
\typeout{** WARNING: IEEEtran.bst: No hyphenation pattern has been}%
\typeout{** loaded for the language `#1'. Using the pattern for}%
\typeout{** the default language instead.}%
\else
\language=\csname l@#1\endcsname
\fi
#2}}
\providecommand{\BIBdecl}{\relax}
\BIBdecl

\bibitem{scellier2017equilibrium}
B.~Scellier and Y.~Bengio, ``Equilibrium propagation: Bridging the gap between energy-based models and backpropagation,'' \emph{Frontiers in computational neuroscience}, vol.~11, p.~24, 2017.

\bibitem{crick1989recent}
F.~Crick, ``The recent excitement about neural networks,'' \emph{Nature}, vol. 337, pp. 129--132, 1989.

\bibitem{martin2021eqspike}
E.~Martin, M.~Ernoult, J.~Laydevant, S.~Li, D.~Querlioz, T.~Petrisor, and J.~Grollier, ``Eqspike: spike-driven equilibrium propagation for neuromorphic implementations,'' \emph{Iscience}, vol.~24, no.~3, 2021.

\bibitem{scellier2019equivalence}
B.~Scellier and Y.~Bengio, ``Equivalence of equilibrium propagation and recurrent backpropagation,'' \emph{Neural computation}, vol.~31, no.~2, pp. 312--329, 2019.

\bibitem{almeida1990learning}
L.~B. Almeida, ``A learning rule for asynchronous perceptrons with feedback in a combinatorial environment,'' in \emph{Artificial neural networks: concept learning}, 1990, pp. 102--111.

\bibitem{pineda1987generalization}
F.~Pineda, ``Generalization of back propagation to recurrent and higher order neural networks,'' in \emph{Neural information processing systems}, 1987.

\bibitem{bi1998synaptic}
G.-q. Bi and M.-m. Poo, ``Synaptic modifications in cultured hippocampal neurons: dependence on spike timing, synaptic strength, and postsynaptic cell type,'' \emph{Journal of neuroscience}, vol.~18, no.~24, pp. 10\,464--10\,472, 1998.

\bibitem{o2019training}
P.~O’Connor, E.~Gavves, and M.~Welling, ``Training a spiking neural network with equilibrium propagation,'' in \emph{The 22nd international conference on artificial intelligence and statistics}.\hskip 1em plus 0.5em minus 0.4em\relax PMLR, 2019, pp. 1516--1523.

\bibitem{ernoult2019updates}
M.~Ernoult, J.~Grollier, D.~Querlioz, Y.~Bengio, and B.~Scellier, ``Updates of equilibrium prop match gradients of backprop through time in an rnn with static input,'' \emph{Advances in neural information processing systems}, vol.~32, 2019.

\bibitem{ernoult2020equilibrium}
------, ``Equilibrium propagation with continual weight updates,'' \emph{arXiv preprint arXiv:2005.04168}, 2020.

\bibitem{laborieux2021scaling}
A.~Laborieux, M.~Ernoult, B.~Scellier, Y.~Bengio, J.~Grollier, and D.~Querlioz, ``Scaling equilibrium propagation to deep convnets by drastically reducing its gradient estimator bias,'' \emph{Frontiers in neuroscience}, vol.~15, p. 633674, 2021.

\bibitem{bal2023equilibrium}
M.~Bal and A.~Sengupta, ``Sequence learning using equilibrium propagation,'' in \emph{2023 32nd International Joint Conference on Artificial Intelligence (IJCAI), arXiv preprint arXiv:2209.09626}, 2023.

\bibitem{mead1990neuromorphic}
C.~Mead, ``Neuromorphic electronic systems,'' \emph{Proceedings of the IEEE}, vol.~78, no.~10, pp. 1629--1636, 1990.

\bibitem{merolla2014million}
P.~A. Merolla, J.~V. Arthur, R.~Alvarez-Icaza, A.~S. Cassidy, J.~Sawada, F.~Akopyan, B.~L. Jackson, N.~Imam, C.~Guo, Y.~Nakamura \emph{et~al.}, ``A million spiking-neuron integrated circuit with a scalable communication network and interface,'' \emph{Science}, vol. 345, no. 6197, pp. 668--673, 2014.

\bibitem{benjamin2014neurogrid}
B.~V. Benjamin, P.~Gao, E.~McQuinn, S.~Choudhary, A.~R. Chandrasekaran, J.-M. Bussat, R.~Alvarez-Icaza, J.~V. Arthur, P.~A. Merolla, and K.~Boahen, ``Neurogrid: A mixed-analog-digital multichip system for large-scale neural simulations,'' \emph{Proceedings of the IEEE}, vol. 102, no.~5, pp. 699--716, 2014.

\bibitem{neftci2019surrogate}
E.~O. Neftci, H.~Mostafa, and F.~Zenke, ``Surrogate gradient learning in spiking neural networks: Bringing the power of gradient-based optimization to spiking neural networks,'' \emph{IEEE Signal Processing Magazine}, vol.~36, no.~6, pp. 51--63, 2019.

\bibitem{he2020comparing}
W.~He, Y.~Wu, L.~Deng, G.~Li, H.~Wang, Y.~Tian, W.~Ding, W.~Wang, and Y.~Xie, ``Comparing snns and rnns on neuromorphic vision datasets: Similarities and differences,'' \emph{Neural Networks}, vol. 132, pp. 108--120, 2020.

\bibitem{yin2021accurate}
B.~Yin, F.~Corradi, and S.~M. Boht{\'e}, ``Accurate and efficient time-domain classification with adaptive spiking recurrent neural networks,'' \emph{Nature Machine Intelligence}, vol.~3, no.~10, pp. 905--913, 2021.

\bibitem{bal2024spikingbert}
M.~Bal and A.~Sengupta, ``Spikingbert: Distilling bert to train spiking language models using implicit differentiation,'' in \emph{Proceedings of the AAAI Conference on Artificial Intelligence}, vol.~38, no.~10, 2024, pp. 10\,998--11\,006.

\bibitem{fang2021deep}
W.~Fang, Z.~Yu, Y.~Chen, T.~Huang, T.~Masquelier, and Y.~Tian, ``Deep residual learning in spiking neural networks,'' \emph{Advances in Neural Information Processing Systems}, vol.~34, pp. 21\,056--21\,069, 2021.

\bibitem{zhou2023spikformer}
\BIBentryALTinterwordspacing
Z.~Zhou, Y.~Zhu, C.~He, Y.~Wang, S.~YAN, Y.~Tian, and L.~Yuan, ``Spikformer: When spiking neural network meets transformer,'' in \emph{The Eleventh International Conference on Learning Representations}, 2023. [Online]. Available: \url{https://openreview.net/forum?id=frE4fUwz_h}
\BIBentrySTDinterwordspacing

\bibitem{hopfield1984neurons}
J.~J. Hopfield, ``Neurons with graded response have collective computational properties like those of two-state neurons.'' \emph{Proceedings of the national academy of sciences}, vol.~81, no.~10, pp. 3088--3092, 1984.

\bibitem{candy1991oversampling}
J.~C. Candy and G.~C. Temes, \emph{Oversampling delta-sigma data converters: theory, design, and simulation}.\hskip 1em plus 0.5em minus 0.4em\relax John Wiley \& Sons, 1991.

\bibitem{he2015delving}
K.~He, X.~Zhang, S.~Ren, and J.~Sun, ``Delving deep into rectifiers: Surpassing human-level performance on imagenet classification,'' in \emph{Proceedings of the IEEE international conference on computer vision}, 2015, pp. 1026--1034.

\bibitem{zenke2021remarkable}
F.~Zenke and T.~P. Vogels, ``The remarkable robustness of surrogate gradient learning for instilling complex function in spiking neural networks,'' \emph{Neural computation}, vol.~33, no.~4, pp. 899--925, 2021.

\end{thebibliography}


\end{document}